\documentclass{jfpc}

\usepackage[pdftex]{graphicx} 
\usepackage{url} 
\usepackage[bookmarks, colorlinks=false, pdfborder={1 1 1}, pdftitle={<pdf title here>}, pdfauthor={<author's name here>}, pdfsubject={<subject here>}, pdfkeywords={<keywords here>}]{hyperref}

\usepackage[linesnumbered,titlenotnumbered,ruled]{algorithm2e}
\usepackage{algorithmic,float}

\usepackage{listings}
\usepackage{geometry}
\usepackage{amssymb}
\usepackage{amsmath}
\usepackage{amssymb}
\usepackage{amsfonts}
\usepackage{csquotes}
\usepackage{ulem}

\newcommand{\qed}{\nobreak \ifvmode \relax \else
      \ifdim\lastskip<1.5em \hskip-\lastskip
      \hskip1.5em plus0em minus0.5em \fi \nobreak
      \vrule height0.75em width0.5em depth0.25em\fi}
\usepackage[table]{xcolor}
\usepackage{longtable}
\usepackage{multirow}
\usepackage{color}
\definecolor{dkgreen}{rgb}{0,0.6,0}
\definecolor{gray}{rgb}{0.5,0.5,0.5}
\definecolor{mauve}{rgb}{0.58,0,0.82}

\annee{2014}
\titre{Une approche CSP pour l'aide à la localisation d'erreurs\footnote{Ce travail a débuté au NII  (National Institute of Informatics, 2-1-2 Hitotsubashi, Chiyoda-ku, Tokyo 101-8430) où Michel  Rueher a été invité plusieurs fois en tant que Professeur associé depuis 2011. Les idées générales et la démarche ont été définies lors des réunions de travail avec les professeurs Hiroshi HOSOBE  et Shin NAKAJIMA.}}

\auteurs{Mohammed Bekkouche \and  Hélène Collavizza \and Michel Rueher} 

\institutions{Univ. Nice Sophia Antipolis, CNRS, I3S, UMR 7271, 06900 Sophia Antipolis, France
}

\mels{\{helen,bekkouche,rueher\}@unice.fr}

\begin{document}
\creationEntete

\begin{resume} 

Nous proposons dans cet article une nouvelle approche basée sur la CP pour l'aide à la localisation des erreurs dans un programme pour lequel un contre-exemple est disponible, c'est à dire que l'on dispose d'une instanciation des variables d'entrée qui viole la post-condition.   Pour aider à localiser les erreurs,  nous générons un système de contraintes pour les chemins du CFG (Graphe de Flot de Contrôle) où au plus k instructions conditionnelles sont susceptibles de contenir  des erreurs. Puis, nous calculons pour chacun de ces chemins  des ensembles minima de correction (ou MCS - Minimal Correction Set) de taille bornée.   Le retrait d'un de ces ensembles de contraintes produit un MSS (Maximal Satisfiable Subset) qui  ne viole plus la post condition.  Nous adaptons pour cela un algorithme proposé par Liffiton et Sakallah \cite{LiS08} afin de pouvoir traiter plus efficacement des programmes avec des calculs numériques. Nous présentons les résultats des premières expérimentations qui sont encourageants.

\end{resume}

\begin{abstract} 

We introduce in this paper a new CP-based approach to support errors location in a program for which a counter-example is available, i.e. an instantiation of the input variables that violates the post-condition. To provide helpful information for error location, we generate a constraint system for the paths of the CFG (Control Flow Graph) for which at most k conditional statements may be erroneous. Then, we calculate Minimal Correction Sets (MCS) of bounded size for each of these paths.  The removal of one of these sets of constraints yields a maximal satisfiable subset,  in other words, a maximal subset of constraints satisfying the post condition. We extend the algorithm proposed by Liffiton and Sakallah \cite{LiS08} to handle programs with numerical statements more efficiently. We present  preliminary experimental results that are quite encouraging.

\end{abstract}

\section{Introduction}
L'aide à la localisation d'erreur à partir de  contre-exemples  ou de traces d'exécution est une question cruciale  lors de la 
mise au point  de logiciels critiques. En effet, quand un programme $P$ contient des erreurs,  un model-checker fournit un 
contre-exemple ou une trace d'exécution qui est souvent longue et difficile à comprendre, et de ce fait d'un intérêt très limité 
pour le programmeur qui doit débugger son programme.  La localisation des portions de code qui contiennent  des erreurs 
est donc souvent un processus difficile et coûteux,  même pour des programmeurs expérimentés. C'est pourquoi nous proposons dans cet 
article une nouvelle approche basée sur la CP pour l'aide à la localisation des erreurs dans un programme pour lequel un 
contre-exemple a été trouvé; c'est à dire pour lequel on dispose d'une instanciation des variables d'entrée qui viole la 
post-condition.   Pour aider à localiser les erreurs,  nous générons un système de contraintes pour les chemins du CFG (Graphe de Flot de Contrôle) où 
au plus $k$ instructions conditionnelles sont susceptibles de contenir  des erreurs. Puis, nous calculons pour chacun de ces 
chemins  des ensembles minima de correction (ou MCS - Minimal Correction Set) de taille bornée.   Le retrait d'un de ces 
ensembles des contraintes initiales produit un MSS (Maximal Satisfiable Subset) qui  ne viole plus la post condition.  Nous adaptons 
pour cela un algorithme proposé par Liffiton et Sakallah afin de pouvoir traiter plus efficacement des programmes avec des calculs 
numériques. Nous présentons les résultats des premières expérimentations qui sont encourageants.

La prochaine section est consacrée à un positionnement de notre approche par rapport aux principales méthodes qui ont été proposées
pour ce résoudre ce problème.  La section suivante est dédiée à la description  de notre approche et des algorithmes utilisés. 
Puis, nous présentons les résultats des premières expérimentations avant de conclure.

\section{\'Etat de l'art}

Dans cette section, nous positionnons notre approche par rapport aux principales méthodes existantes. Nous parlerons d'abord des 
méthodes utilisées pour l'aide à la localisation des erreurs dans le cadre du test et de la vérification de programmes. Comme 
l'approche que nous proposons consiste essentiellement à rechercher des sous-ensembles de contraintes spécifiques  dans un système 
de contraintes inconsistant, nous parlerons aussi dans un second temps  des algorithmes qui ont été utilisés en recherche 
opérationnelle et en programmation par contraintes pour aider l'utilisateur à debugger un système de contraintes inconsistant.

\subsection{Méthodes d'aide à la localisation des erreurs utilisées en test et en vérification de programmes}

Différentes approches ont été proposées pour aider le programmeur dans l'aide à la localisation d'erreurs dans la communauté test et 
vérification. 

Le problème de la localisation des erreurs a d'abord été abordé dans le cadre du test où de nombreux systèmes ont été développés. Le plus célèbre est Tarantula \cite{JonesHS02, JonesH05} qui utilise différentes métriques pour établir un classement des instructions suspectes détectées lors de l'exécution d'une batterie de tests. Le point critique de cette approche réside dans le fait qu'elle requiert un oracle qui permet de décider si le résultat du test est juste ou non.
Nous nous plaçons ici dans un cadre moins exigeant (et plus réaliste) qui est celui du Bounded-Model Checking (BMC), c'est à dire un cadre où les seuls prérequis sont un programme, une post-condition ou une assertion qui doit être vérifiée, et éventuellement une pré-condition.

C'est aussi dans ce cadre que se placent Bal et al\cite {BNR03} qui utilisent plusieurs appels à un Model Checker et comparent les contre-exemples obtenus avec une  trace d'exécution correcte. Les transitions qui ne figurent pas dans la
trace correcte sont signalées comme une possible cause de l'erreur. Ils ont implanté leur algorithme dans le contexte de SLAM, 
un model checker qui vérifie les propriétés de sécurité temporelles de programmes C.

Plus récemment, des approches basées sur la dérivation de traces correctes ont été introduites dans un système nommé {\tt Explain}\cite{GKL04,GCK06} et qui
fonctionne en trois étapes :
\begin{enumerate}
\item  Appel de {\tt CBMC }\footnote{\url{http://www.cprover.org/cbmc/}} pour trouver une exécution  qui viole la post-condition;
\item  Utilisation d'un solveur pseudo-booléen pour rechercher l'exécution correcte la plus proche;
\item  Calcul de la différence entre les traces. 
\end{enumerate}
{\tt Explain} produit ensuite  une formule propositionnelle $S$ associée à {\tt P} 
mais dont les affectations ne violent pas la spécification.  Enfin {\tt Explain} étend  $S$ avec des contraintes représentant un
 problème d'optimisation : trouver une affectation satisfaisante qui soit aussi
proche que possible du contre-exemple; la proximité étant mesurée  par une  distance sur les exécutions de {\tt P}. 

Une approche similaire à {\tt Explain} a été introduite dans  \cite{ReR03} mais elle
est basée sur le test plutôt que sur la vérification de modèles : les auteurs utilisent des séries de tests correctes  et des séries erronées. 
Ils utilisent aussi des métriques de distance pour sélectionner un test correct à partir d'un ensemble donné de tests. Cette 
approche suppose  qu'un oracle soit disponible.

Dans  \cite{GBC06, GSB07}, les auteurs partent aussi de la trace d'un contre-exemple, mais ils utilisent la spécification pour dériver un  programme 
correct pour les mêmes données d'entrée.  Chaque instruction identifiée est un candidat potentiel de faute et elle peut être utilisée pour corriger les
 erreurs. Cette approche garantit que les erreurs sont effectivement parmi les instructions identifiées (en supposant que l'erreur est dans le modèle 
erroné considéré). En d'autres termes, leur approche identifie un sur-ensemble des instructions  erronées. Pour réduire le nombre d'erreurs 
potentielles, le processus est redémarré pour différents contre-exemples et les auteurs calculent l'intersection des ensembles d'instructions 
suspectes. Cependant, cette approche souffre de deux problèmes majeurs: 
\begin{itemize}
\item Elle permet de modifier n'importe quelle expression, l'espace de recherche peut ainsi être très grand ;
\item  Elle peut renvoyer beaucoup de faux diagnostics totalement absurdes car toute modification d'expression est possible (par exemple, changer la 
dernière affectation d'une fonction pour renvoyer le résultat attendu).
\end{itemize}

Pour remédier à ces inconvénients, Zhang et al \cite{ZGG06} proposent de modifier uniquement les prédicats de flux  de contrôle.  
L'intuition de cette approche   est qu'à travers un switch  des résultats d'un prédicat et la modification du flot de contrôle, 
l'état de programme peut non seulement être modifié à peu de frais, mais qu'en plus, il est souvent possible d'arriver à un état succès. 
Liu et al \cite{LiL10} généralisent cette approche en permettant la modification de plusieurs prédicats. Ils proposent également une étude théorique 
d'un algorithme de débogage pour les erreurs {\it RHS}, c'est  à dire les erreurs dans les prédicats de contrôle et la partie droite des affectations. 

Dans \cite{BBCOT09}, les auteurs abordent le problème de l'analyse de la trace d'un  contre-exemple et de l'identification de l'erreur dans le cadre des systèmes de 
vérification formelle du hardware. Ils utilisent pour cela  la notion de causalité introduite par Halpern et Pearl pour définir formellement une 
série de causes de la violation de la spécification par un contre-exemple.

Récemment, Manu Jose et Rupak Majumdar \cite{JoM11a,JoM11b} ont abordé ce problème différemment : ils ont introduit un nouvel algorithme qui utilise un solveur MAX-SAT pour 
calculer le nombre maximum des clauses d'une formule booléenne qui peut être satisfaite par une affectation. Leur algorithme fonctionne en trois étapes: 
\begin{enumerate}
\item ils encodent une trace d'un programme par une formule booléenne $F$ qui est satisfiable si et seulement si 
 la trace est satisfiable ; 
\item ils construisent une formule fausse $F '$ en imposant que la post-condition soit  vraie (la formule $F '$  est  insatisfiable car la 
trace correspond à un contre-exemple qui viole la post-condition) ;
\item Ils utilisent {\tt MAX\-SAT } pour calculer le nombre maximum de clauses
pouvant être satisfaites dans $F '$ et affichent le complément de cet ensemble comme une cause potentielle des erreurs. En d'autres termes, ils calculent le complément d'un MSS (Maximal Satisfiable Subset).
\end{enumerate}
Manu Jose et Rupak Majumdar~\cite{JoM11a,JoM11b} ont implanté  leur algorithme dans un outil appelé {\tt Bug\-Assist} qui utilise {\tt CBMC}.

Si-Mohamed Lamraoui et  Shin Nakajima~\cite{SiN14} ont aussi développé récemment un outil nommé  {\tt SNIPER} qui calcule les MSS  d'une formule $\psi = EI \wedge TF \wedge AS$ où $EI$ encode les valeurs d'entrée erronées, $TF$ est une formule qui représente tous les chemins du programme, et $AS$ correspond à l'assertion qui est violée.  Les MCS sont obtenus en prenant le complément des MSS calculés. L'implémentation est basée sur la représentation  intermédiaire  {\tt LLVM} et le solveur SMT {\tt Yices}. L'implémentation actuelle est toutefois beaucoup plus lente que  {\tt Bug\-Assist}.

 L'approche que nous proposons ici est inspirée des 
travaux de Manu Jose et Rupak Majumdar. Les principales différences sont:
\begin {enumerate}
\item Nous ne transformons pas tout le programme en un système de contraintes mais nous utilisons le graphe de flot de contrôle pour collecter les contraintes du chemin du contre exemple et  des chemins dérivés de ce dernier en supposant qu'au plus k instructions conditionnelles sont susceptibles de contenir  des erreurs.
\item Nous n'utilisons pas des algorithmes basés sur {\tt MAX\-SAT } mais des algorithmes plus généraux qui permettent plus facilement de traiter des contraintes numériques.
\end {enumerate}

\subsection{Méthodes pour débugger un système de contraintes inconsistant}

En recherche  opérationnelle et en programmation par contraintes, différents algorithmes  ont été proposés pour aider l'utilisateur 
à debugger un 
système de contraintes inconsistant. 
Lorsqu'on recherche des informations utiles pour la localisation des erreurs sur les systèmes de contraintes numériques, on peut s'intéresser à deux types d'informations:
\begin{enumerate}
\item Combien de contraintes dans un ensemble de contraintes insatisfiables peuvent être satisfaites? 
\item Où  se situe le problème  dans  le système de contraintes ?
\end{enumerate}
Avant de présenter rapidement les algorithmes\footnote{Pour une présentation plus détaillée voir  
\url{ http://users.polytech.unice.fr/~rueher/Publis/Talk_NII_2013-11-06.pdf}} qui cherchent à répondre à ces questions, nous allons définir plus formellement les notion de MUS, MSS et MCS à l'aide des  définitions introduites  dans \cite{LiS08}.\\
Un MUS (Minimal Unsatisfiable Subsets) est un ensemble de contraintes qui est inconsistant mais qui devient consistant si une contrainte quelconque est retirée de cet ensemble. Plus formellement, soit $C$ un ensemble de contraintes:  \\
\centerline{$M \subseteq C$ est un MUS $ \Leftrightarrow  \, M$ est UNSAT}\\
\centerline{et $\forall c \in M: M \setminus \{c\}$ est SAT.}\\
La notion de MSS (Maximal Satisfiable Subset) est une généralisation de MaxSAT / MaxCSP où l'on considère la maximalité au lieu de 
la cardinalité maximale:\\
\centerline{$M \subseteq C$ est un MSS $ \Leftrightarrow  \, M$ est SAT}\\
\centerline{et $\forall c \in C \setminus M : M \cup \{c\}$ est UNSAT.}\\
Cette définition est très proche de celle des IIS (Irreducible Inconsistent Subsystem) utilisés en recherche 
opérationnelle~\cite{Chin96,Chin01,Chin08}.\\
Les MCS (Minimal Correction Set) sont des compléments des MSS (le retrait d'un MCS à $C$ produit un MSS car on ``corrige'' l'infaisabilité): \\
\centerline{$M \subseteq C$ est un MCS $ \Leftrightarrow  \, C  \setminus M$ est SAT}\\
\centerline{et $\forall c \in M : (C \setminus M) \cup \{c\}$ est UNSAT.}\\
Il existe donc une dualité entre l'ensemble des MUS et des MCS \cite{Bil03,LiS08} : informellement,  l'ensemble des MCS est équivalent aux ensembles couvrants 
irréductibles\footnote{Soit $\Sigma$  un ensemble d'ensemble et $D$ l'union des élements de  $\Sigma$.  On rappelle que $H$ est un ensemble couvrant  de $\Sigma$ si $H \subseteq D$ et  $\forall S \in \Sigma : H \cup S \neq \emptyset$. $H$ est irréductible (ou minimal) si aucun élément ne peut être retiré de $H$  sans que celui-ci ne perde sa propriété d'ensemble couvrant.}\\ des MUS; et l'ensemble des MUS est équivalent aux ensembles couvrants irréductibles des MCS. 
Soit un ensemble de contraintes $C$:
\begin{enumerate}
\item Un sous-ensemble $M$ de $C$ est un MCS ssi M est un ensemble couvrant minimal des MUS de $C$;
\item Un sous-ensemble $M$ de $C$ est un MUS ssi M est un ensemble couvrant minimal des MCS de $C$;
\end{enumerate}
Au niveau intuitif, il est aisé de comprendre qu'un MCS doit au moins retirer une contrainte de chaque MUS. Et comme un MUS peut être rendu satisfiable en retirant n'importe laquelle de ses contraintes,  chaque MCS doit au moins contenir une contrainte de chaque MUS.
Cette dualité est aussi intéressante pour notre problématique car elle montre que les réponses aux deux questions posées ci-dessus sont étroitement liées. \

Différents algorithmes ont été proposés pour le calcul des IIS/MUS et MCS. Parmi les premiers travaux, ont peut mentionner les algorithmes {\tt Deletion Filter},  {\tt Additive Method},  {\tt Additive Deletion Method},  {\tt Elastic Filter} qui ont été développés dans la communauté  de recherche opérationnelle~\cite{Chin96,TMJ96, Chin01, Chin08}. Les trois premiers algorithmes sont des algorithmes itératifs alors que le quatrième utilise des variables d'écart pour identifier dans la première phase du Simplexe  les  contraintes susceptibles de figurer dans un IIS.\\
Junker \cite{Jun04} a proposé un algorithme générique basé sur une stratégie "Divide-and-Conquer" pour calculer efficacement les IIS/MUS lorsque la taille des sous-ensembles conflictuels est beaucoup plus petite que celle de l'ensemble total des contraintes.\\
L'algorithme de Liffiton et Sakallah~\cite{LiS08} qui calcule d'abord l'ensemble des MCS  par ordre de taille croissante, puis l'ensemble des MUS est basé sur la propriété mentionnée ci-dessus. Cet algorithme, que nous avons utilisé dans notre implémentation est décrit dans la section suivante.

Différentes améliorations~\cite{FeSZ12,LiM13,MHJPB13} de ces algorithmes ont été proposées ces dernières années  mais elles sont assez étroitement liées à SAT et a priori assez difficilement transposables dans un contexte où nous avons de nombreuses contraintes numériques.

\section{Notre approche}

Dans cette section nous allons d'abord présenter le cadre général de notre approche, à savoir celui du ``Bounded Model Checking'' (BMC) basé sur la programmation par
 contraintes, puis nous allons décrire la méthode proposée et les algorithmes utilisés pour calculer des MCS de cardinalité bornée.

\subsection{Les principes : BMC et MCS }

Notre approche se place dans le cadre du ``Bounded model Checking'' (BMC) par programmation par contraintes \cite{CRH10,CVPRR14}. En BMC, les programmes sont dépliés en utilisant une borne $b$, c'est à dire que les boucles sont remplacées par des imbrications de conditionnelles de profondeur au plus $b$. Il s'agit ensuite de détecter des non-conformités par rapport à une spécification. \'Etant donné un triplet de Hoare \{$PRE$,$PROG_b$,$POST$\}, où $PRE$ est la pré-condition, $PROG_b$ est le programme déplié $b$ fois et $POST$ est la post-condition, le programme est {\it non conforme} si la formule $\Phi = PRE \wedge PROG_b \wedge \neg POST$ est satisfiable. Dans ce cas, une instanciation des variables de $\Phi$ est un {\it contre-exemple}, et un cas de non conformité, puisqu'il satisfait à la fois la pré-condition et le programme, mais ne satisfait pas la post-condition. 

{\tt CPBPV} \cite{CRH10} est un outil de BMC  basé sur la programmation par contraintes. $CPBPV$ transforme $PRE$ et $POST$ en contraintes, et transforme $PROG_b$ en un CFG dans lequel les conditions et les affectations sont  traduites en contraintes\footnote{Pour éviter les problèmes de re-définitions multiples des variables, la forme DSA (Dynamic Single Assignment~\cite{BarnettLeino05}) est utilisée}. {\tt CPBPV}  construit le {\tt CSP} de la formule $\Phi$ {\it à la volée}, par un parcours en profondeur du graphe. 
À l'état initial, le {\tt CSP} contient les contraintes de $PRE$ et $\neg POST$, puis les contraintes d'un chemin sont ajoutées au fur et à mesure de l'exploration du graphe. Quand le dernier noeud d'un chemin est atteint, la faisabilité du {\tt CSP} est testée. S'il est consistant, alors on a trouvé un contre-exemple, sinon, un retour arrière est effectué pour explorer  une autre branche du CFG. Si tous les chemins ont été explorés sans trouver de contre-exemple, alors le programme est conforme à sa spécification (sous réserve de l'hypothèse de dépliage des boucles).

Les travaux présentés dans cet article cherchent à {\it localiser} l'erreur détectée par la phase de BMC. Plus précisemment, soit $CE$ une instanciation des variables qui satisfait le {\tt CSP} contenant les contraintes de $PRE$ et $\neg POST$, et les contraintes  d'un chemin incorrect  de $PROG_b$ noté $PATH$. Alors le {\tt CSP}
 $C = CE \cup PRE \cup PATH \cup POST$ est {\it inconsistant}, puisque  $CE$ est un contre-exemple et ne satisfait donc pas la post-condition. Un {\it ensemble minima de correction} (ou MCS - Minimal Correction Set) de $C$ est un ensemble de contraintes qu'il faut nécessairement enlever pour que $C$ devienne consistant. 
Un tel ensemble fournit donc une {\it localisation de l'erreur} sur le chemin du contre-exemple. Comme l'erreur peut se trouver dans une affectation sur le chemin du contre-exemple, mais peut aussi provenir d'un mauvais branchement, notre approche (nommée  {\tt LocFaults}) s'intéresse également aux MCS des systèmes de contraintes obtenus en déviant des branchements par rapport au comportement induit par le contre-exemple. Plus précisément, l'algorithme $LocFaults$  effectue un parcours en profondeur d'abord du $CFG$ de $PROG_b$, en propageant le contre-exemple et en déviant au plus $k_{max}$ conditions. Trois cas peuvent être distingués:
\begin{itemize}
\item Aucune condition n'a été déviée:  {\tt LocFaults} a parcouru le chemin du contre-exemple en collectant les contraintes de ce chemin et il va calculer les MCS sur cet ensemble de contraintes;
\item  $k_{max}$ conditions ont été déviées sans qu'on arrive à trouver un chemin qui satisfasse la post-condition: on abandonne l'exploration  de ce chemin;
\item $d$ conditions  on déja été déviées et on peut encore dévier au moins une condition,  c'est à dire  $k_{max}\ > 1$. Alors la  condition courante $c$ est déviée.  Si le chemin résultant ne viole plus la post-condition, l'erreur sur le chemin initial peut avoir deux causes différentes :
\begin{itemize}
\item[(i)]  les conditions déviées elles-mêmes sont cause de l'erreur,
\item[(ii)]  une erreur dans une affectation a provoqué une mauvaise évaluation de $c$, faisant  prendre le mauvais branchement. 
\end{itemize}
Dans le cas (ii), le  {\tt CSP} $CE \cup PRE \cup PATH_{c} \cup  \{c\}$, où $PATH_{c}$ est l'ensemble des contraintes du chemin courant, c'est à dire le chemin du contre-exemple dans lequel $d$ déviations ont déjà été prises, est satisfiable. 
Par conséquent, le {\tt CSP} $ CE \cup PRE \cup PATH_{c} \cup {\{\textcolor{red}\neg} c\}$ est {\it insatisfiable}. {\tt LocFaults} calcule donc également les MCS de ce {\tt CSP}, afin de détecter les instructions suspectées d'avoir induit le mauvais branchement pour $c$.

Bien entendu, nous ne calculons pas les MCS des chemins ayant le même préfixe : si la déviation d'une condition permet de satisfaire  la post-condition, il est inutile de chercher à modifier des conditions supplémentaires dans la suite du chemin

\end{itemize}

\subsection{Description de l'algorithme}

\begin{algorithm*}[!Htf]

\caption{LocFaults}
\label{LocFaults}
\algsetup{linenosize=\tiny,linenodelimiter=.,indent=2em}
{\fontsize{8}{8}\selectfont

{\tt Fonction LocFaults}($PROG_b$,$CE$,$k_{max}$,$MCS_b$)\\
\Entree{
\begin{itemize}
\item $PROG_b$: un programme déplié b fois non conforme vis-à-vis de sa spécification, 
\item $CE$: un contre-exemple de $PROG_b$,
\item $k_{max}$ : le nombre maximum de conditions à dévier,
\item $MCS_b$: la borne du cardinal des MCS
\end{itemize}}
\Sortie{une liste de corrections possibles}
\Deb{
$CFG$ $\leftarrow$ $CFG\_build(PROG_b)$ \% construction du CFG \\
$MCS$ = [] \\
$DFS_{devie}$($CFG.root$,$CE$,$\emptyset$,$\emptyset$,$0$,$MCS$,$MCS_b$) \% calcul des MCS sur le chemin du contre-exemple \\
$DFS_{devie}$($CFG.root$,$CE$,$\emptyset$,$\emptyset$,$k_{max}$,$MCS$,$MCS_b$) \% calcul des MCS en prenant au plus $k_{max}$  déviations\\
\Retour $MCS$
}
}
\end{algorithm*}

\begin{algorithm*}[!Htf]

\caption{$DFS_{devie}$}
\label{dfs_devie}
\algsetup{linenosize=\tiny,linenodelimiter=.,indent=2em}
{\fontsize{8}{8}\selectfont

{\tt Fonction $DFS_{devie}$}($n$,$P$,$CSP_d$,$CSP_a$,$k$,$MCS$,$MCS_b$)\\
\Entree{
\begin{itemize}
\item $n$: noeud du CFG, 
\item $P$: contraintes de propagation (issues du contre-exemple et du chemin), 
\item $CSP_d$: contraintes des conditions déviées, 
\item $CSP_a$: contraintes des affectations, 
\item $k$: nombre de conditions à dévier, 
\item $MCS$: ensemble des MCS calculés, 
\item $MCS_b$: la borne du cardinal des MCS
\end{itemize}}
\Deb{
\Si{$n$ est la postcondition} { \% on est sur le chemin du CE, calcul des MCS  \\
$CSP_a$ $\leftarrow$ $CSP_a$ $\cup$ $\{cstr(POST)\}$ \\
 $MCS.add(MCS(CSP_a,MCS_b))$
}

\SinonSi{$n$ est un noeud conditionnel}{ 
\Si{$P$ $\cup$ \{$cstr(n.cond)$\} est faisable}{ \% $next$ est le noeud où l'on doit aller, $devie$ est la branche opposée \\
$next=n.gauche$ \\ 
$devie=n.droite$
}
\Sinon{
$next=n.droite$ \\
$devie=n.gauche$
}
\Si{$k > 0$} {
\% on essaie de dévier la condition courante\\
corrige = {\tt correct}($devie$,$P$) \\
\Si{corrige}{ \% le chemin est corrigé, on met à jour les MCS \\
$CSP_d \leftarrow CSP_d$ $\cup$ \{$cstr(n.cond$)\} \\
$MCS.addAll(CSP_d)$  \% ajout des conditions déviées \\
\% calcul des MCS sur le chemin qui m\`ene à la dernière condition déviée  \\
\PourCh{ c dans $CSP_d$ } {   
$CSP_a$ $\leftarrow$ $CSP_a$ $\cup$ \{$\neg c$\}  
}
  $MCS.add(MCS(CSP_a,MCS_b))$  \\
}
\SinonSi{k>1}{
\% on essaie de dévier la condition courante et des conditions en dessous \\
$DFS_{devie}$($devie$,$P$,$CSP_d$ $\cup$ $\{cstr(n.cond)\}$,$CSP_a$,$k-1$,$MCS$,$MCS_b$)}

\% dans tous les cas, on essaie de dévier les conditions en dessous du noeud courant \\
$DFS_{devie}$($next$,$P$,$CSP_d$,$CSP_a$,$k$,$MCS$,$MCS_b$) 
}

\Sinon{ \% k=0, on est sur le chemin du contre-exemple, on suit le chemin \\
$DFS_{devie}$($next$,$P$,$CSP_d$,$CSP_a$,$k$,$MCS$,$MCS_b$)}
}

\SinonSi{($n$ est un bloc d'affectations)}{
\PourCh{ affectation $ass$ $\in$ $n.assigns$}{  
P.add(propagate(ass,$P$))   \\
$CSP_a$ $\leftarrow$ $CSP_a$ $\cup$ $\{cstr(ass)\}$  
}
\% On continue l'exploration sur le noeud suivant  \\
$DFS_{devie}$($n.next$,$CE$,$P$,$CSP_d$,$CSP_a$,$k$,$MCS$,$MCS_b$)  
}
}
}
\end{algorithm*} 

\begin{algorithm}[!Htf]

\caption{correct}
\label{correct}
\algsetup{linenosize=\tiny,linenodelimiter=.,indent=2em}
{\fontsize{8}{8}\selectfont
  
{\tt Fonction correct}($n$,$P$)\\
\Entree{
\begin{itemize}
\item $n$: noeud du CFG, 
\item $P$: contraintes de propagation (issues du contre-exemple et du chemin), 
\end{itemize}}
\Sortie{true si le programme est correct sur le chemin induit par $P$}

\Deb{
\Si{$n$ est la postcondition} {

\Si{$P$ $\cup$ $\{cstr(POST)\}$ est faisable} { 
\Retour true }
\Sinon{ \Retour false }
}

\SinonSi{$n$ est un noeud conditionnel}{  
\Si{$P$ $\cup$ \{$cstr(n.cond)$\} est faisable}{ \%  exploration de la branche If  \\
\Retour correct($n.left$,$P$)  
}
\Sinon {\Retour correct($n.right$,$P$)  }

}

\SinonSi{($n$ est un bloc d'affectations)}{  
\% on propage les affectations \\
\PourCh{ affectation $ass$ $\in$ $n.assigns$}{  
 P.add(propagate(ass,$P$))
}
\% On continue l'exploration sur le noeud suivant  \\
\Retour correct($n.next$,$P$,$CSP_d$,$CSP_a$,$MCS$,$MCS_b$)  
}
}
}

\end{algorithm}

\begin{algorithm}[!Htf]
 
\caption{MCS}
\label{MCS} 
{\tt Fonction MCS}($C$,$MCS_b$)\\
\Entree{$C$: Ensemble de contraintes infaisable, $MCS_b$: Entier}
\Sortie{$MCS$: Liste de MCS de $C$ de cardinalité inférieure à $MCS_b$}
\Deb{
$C'$ $\leftarrow$ \textsc{AddYVars}($C$) \\
$MCS$ $\leftarrow$ $\emptyset$ \\
$k$ $\leftarrow$ $1$ \\ 
\Tq{\textsc{SAT}($C'$) $\land$ $k$<=$MCS_b$}{
$C'_{k}$ $\leftarrow$ $C'$ $\land$ \textsc{AtMost}($\{\neg y_{1},\neg y_{2},...,\neg y_{n}\}$,$k$)\\
\Tq{SAT($C'_{k}$)}{
$MCS.add(newMCS)$.\\ 
$C'_{k}$ $\leftarrow$ $C'_{k}$ $\land$ \textsc{BlockingClause}($newMCS$) 
\\
$C'$ $\leftarrow$ $C'$ $\land$ \textsc{BlockingClause}($newMCS$). 
}
$k$ $\leftarrow$ $k$ + $1$.\\
}
\Retour $MCS$
}

\end{algorithm}

L'algorithme {\tt LocFaults} (cf. Algorithm \ref{LocFaults}) prend en entrée un programme déplié non conforme vis-à-vis de sa spécification,
 un contre-exemple, et une borne maximum de la taille des ensembles de correction. Il dévie au plus $k_{max}$ conditions par rapport au contre-exemple fourni,  et renvoie une liste de corrections possibles. 

{\tt LocFaults} commence par construire le CFG du programme  puis appelle la fonction {\tt DFS} sur le chemin du contre-exemple (i.e. en déviant $0$ condition) puis en acceptant au plus $k_{max}$ déviations.  La fonction {\tt DFS} g\`ere trois ensembles de contraintes : 
\begin{itemize}
\item $CSP_d$: l'ensemble des contraintes des conditions qui ont été déviées à partir du chemin du contre-exemple,
\item $CSP_a$: l'ensemble des contraintes d'affectations du chemin,
\item $P$: l'ensemble des contraintes dites de propagation, c'est à dire les contraintes de la forme $variable = constante$ qui sont obtenues en propageant le contre exemple sur les affectations du chemin. 
\end{itemize}
L'ensemble $P$ est utilisé pour propager les informations et vérifier si une condition est satisfaite, les ensembles $CSP_d$ et $CSP_a$ sont utilisés pour calculer les $MCS$. Ces trois ensembles sont collectés à la volée lors du parcours en profondeur.
Les paramètres de la fonction {\tt DFS} sont les ensembles $CSP_d$, $CSP_a$  et $P$ décrits ci-dessus, $n$ le noeud courant  du CFG,   $MCS$ la liste 
des corrections en cours de construction, $k$ le nombre de déviations autorisées et $MCS_b$ la borne de la taille des $MCS$. Nous notons $n.left$ (resp. $n.right$) la branche {\it if} (resp. {\it else}) d'un noeud conditionnel, et $n.next$  le noeud qui suit un bloc d'affectation; $cstr$ est la fonction qui traduit une condition ou affectation en contraintes.

Le parcours commence avec $CSP_d$ et $CSP_a$ vides et $P$ contenant les contraintes du contre-exemple. Il part de la racine de l'arbre ($CFG.root$) qui contient la pré-condition, et se déroule comme suit :
\begin{itemize}
\item Quand le dernier noeud est atteint (i.e. noeud de la post-condition), on est sur le chemin du contre-exemple. La post-condition est ajoutée à $CSP_a$ et on cherche les MCS,

\item Quand le noeud est un noeud conditionnel, alors on utilise $P$ pour savoir si la condition est satisfaite. Si on peut encore prendre une déviation (i.e. $k>0$), on essaie de dévier la condition courante $c$ et on vérifie si cette déviation corrige le programme en appelant la fonction $correct$. Cette fonction propage tout simplement le contre-exemple sur le graphe à partir du noeud courant et renvoie vrai si le programme satisfait la  postcondition  pour ce chemin,
\begin{itemize}
\item Si dévier $c$ a corrigé le programme, alors les conditions qui ont été déviées (i.e. $CSP_d \cup c$) sont des corrections. De plus, on calcule aussi les corrections dans le chemin menant à $c$,
\item Si dévier $c$ n'a pas corrigé le programme, si on peut encore dévier des conditions (i.e. $k\geq1$) alors on dévie $c$ et on essaie de  dévier $k-1$ conditions en dessous de $c$.
\end{itemize}
Dans les deux cas (dévier $c$ a corrigé ou non le programme), on essaie aussi de dévier des conditions en dessous de $c$, sans dévier $c$.

\item Quand le noeud est un bloc d'affectations, on propage le contre-exemple sur ces affectations et on ajoute les contraintes correspondantes dans $P$ et dans $CSP_a$.
\end{itemize}

L'algorithme {\tt LocFaults} appelle l'algorithme {\tt MCS} (cf. Algorithm \ref{MCS}) qui est une 
transcription directe de l'algorithme proposé par Liffiton et Sakallah \cite{LiS08}. Cet algorithme associe à chaque contrainte un sélecteur de variable $y_i$ qui peut prendre la valeur 0 ou 1; la contrainte {\tt AtMost} permet donc de retenir au plus $k$ contraintes du système de contraintes initial dans le MCS.
La procédure {\tt BlockingClause(newMCS)} appelée à la ligne  10 (resp. ligne 11 ) permet d'exclure les sur-ensembles de taille $k$ (resp. de taille supérieure à $k$). 

Lors de l'implémentation de cet algorithme nous avons utilisé~{\tt IBM ILOG CPLEX}\footnote{\url{http://www-01.ibm.com/software/commerce/optimization/cplex-optimizer/}} qui permet à la fois une implémentation aisée de la fonction  {\tt AtMost} et la résolution de systèmes de contraintes numériques. Il faut toutefois noter que cette résolution n'est correcte que sur les entiers et que la prise en compte des nombres flottants nécessite l'utilisation d'un solveur spécialisé pour le traitement des flottants.

\section{Evaluation expérimentale}
\begin{table*}[!Htf]
\begin{center}
\begin{tiny}
\begin{tabular}{|c|c|c|c|c|c|c|c|}
\hline
\multirow{2}{*}{Programme} & \multirow{2}{*}{Contre-exemple} & \multirow{2}{*}{Erreurs} & \multicolumn{4}{|c|}{LocFaults} & \multirow{2}{*}{BugAssist} \\ 
\cline{4-7}
  &   &   & $= 0$ & $\leq 1$ & $\leq 2$ & $\leq 3$ & \\
\hline
{\tiny AbsMinusKO}  & $\{i=0,j=1\}$  & $17$  &  $\{\textcolor{red}{17}\}$  &  $\{\textcolor{red}{17}\}$  &  $\{\textcolor{red}{17}\}$  & $\{\textcolor{red}{17}\}$  & $\{\textcolor{red}{17}\}$ \\ 
\hline
{\tiny AbsMinusKO2}  &  $\{i=0,j=1\}$  & $11$  &  $\{\textcolor{red}{11}\}$,$\{17\}$  &  $\{\textcolor{red}{11}\}$,$\{17\}$  &  $\{\textcolor{red}{11}\}$,$\{17\}$   &  $\{\textcolor{red}{11}\}$,$\{17\}$  &  $\{17,20,16\}$ \\
\hline 
\multirow{2}{*}{{\tiny AbsMinusKO3}}  & \multirow{2}{*}{$\{i=0,j=1\}$}  & \multirow{2}{*}{$14$}  &  \multirow{2}{*}{$\{20\}$}  &  $\{\uline{16}\}$,$\{\textcolor{red}{14}\}$,$\{12\}$,  &  $\{\uline{16}\}$,$\{\textcolor{red}{14}\}$,$\{12\}$,  &  $\{\uline{16}\}$,$\{\textcolor{red}{14}\}$,$\{12\}$,  &  \multirow{2}{*}{$\{16,20\}$} \\
  &   &   &  & $\{20\}$ & $\{20\}$ & $\{20\}$ & \\
\hline
\multirow{2}{*}{{\tiny MinmaxKO}}  &  $\{in_{1}=2,in_{2}=1,$  &  \multirow{2}{*}{$19$}  &  \multirow{2}{*}{$\{10\}$,$\{\textcolor{red}{19}\}$}  &  $\{\uline{18}\}$,$\{10\}$    &   $\{\uline{18}\}$,$\{10\}$  &  $\{\uline{18}\}$,$\{10\}$  & \multirow{2}{*}{$\{18,\textcolor{red}{19},22\}$} \\
\cline{5-7}
  & $in_{3}=3\}$  &   &   &  $\{10\}$,$\{\textcolor{red}{19}\}$ & $\{10\}$,$\{\textcolor{red}{19}\}$  &  $\{10\}$,$\{\textcolor{red}{19}\}$  &   \\
\hline
\multirow{2}{*}{{\tiny MidKO}}   &  \multirow{2}{*}{$\{a=2,b=1,c=3\}$}  & \multirow{2}{*}{$19$} &  \multirow{2}{*}{$\{\textcolor{red}{19}\}$}  &  \multirow{2}{*}{$\{\textcolor{red}{19}\}$}  &  \multirow{2}{*}{$\{\textcolor{red}{19}\}$}  & $\{\uline{14},\uline{23},\uline{26}\}$  & \multirow{2}{*}{$\{14,\textcolor{red}{19},30\}$} \\
\cline{7-7}
&  &  &  &  &   & $\{\textcolor{red}{19}\}$  &  \\
\hline
\multirow{3}{*}{{\tiny Maxmin6varKO}}   &  \multirow{2}{*}{$\{a=1,b=-4,c=-3,$}  &  \multirow{3}{*}{$27$}  &  \multirow{3}{*}{$\{28\}$}   &   $\{\uline{15}\}$  &  $\{\uline{15}\}$  & $\{\uline{15}\}$  & \multirow{2}{*}{$\{15,12,\textcolor{red}{27},$}  \\ 
\cline{5-7} 
               &   \multirow{2}{*}{$d=-1,e=0,f=-4\}$}  &    &     &  $\{\uline{\textcolor{red}{27}}\}$   &  $\{\uline{\textcolor{red}{27}}\}$  &  $\{\uline{\textcolor{red}{27}}\}$ & \multirow{2}{*}{$31,166\}$}  \\    
\cline{5-7}               
&  &  &  & $\{28\}$ & $\{28\}$  &  $\{28\}$ &  \\               
\hline
\multirow{2}{*}{{\tiny Maxmin6varKO2}}   & $\{a=1,b=-3,c=0,$   &  \multirow{2}{*}{$12$}  &  \multirow{2}{*}{$\{65\}$}   &  $\{\uline{\textcolor{red}{12}}\}$   &  $\{\uline{\textcolor{red}{12}}\}$  & $\{\uline{\textcolor{red}{12}}\}$  & \multirow{2}{*}{$\{\textcolor{red}{12},64,166\}$}  \\
\cline{5-7}
               &   $d=-2,e=-1,f=-2\}$  &    &     &  $\{65\}$   &  $\{65\}$  & $\{65\}$ &   \\    
\hline
\multirow{2}{*}{{\tiny Maxmin6varKO3}}   & $\{a=1,b=-3,c=0,$  &  \multirow{2}{*}{$12$,$15$}  &  \multirow{2}{*}{$\{65\}$}  &  \multirow{2}{*}{$\{65\}$}  &  $\{\uline{\textcolor{red}{12}},\uline{\textcolor{red}{15}}\}$  & $\{\uline{\textcolor{red}{12}},\uline{\textcolor{red}{15}}\}$  &  $\{\textcolor{red}{12},\textcolor{red}{15},64,$ \\
\cline{6-7}
               &   $d=-2,e=-1,f=-2\}$  &    &     &     &  $\{65\}$  &  $\{65\}$  &  $166\}$ \\    
\hline
\multirow{2}{*}{{\tiny Maxmin6varKO4} }  &  $\{a=1,b=-3,c=-4,$  &  $12$,$15$,&  \multirow{2}{*}{$\{116\}$}   &  \multirow{2}{*}{$\{116\}$}   &  \multirow{2}{*}{$\{116\}$}  & $\{\uline{\textcolor{red}{12}},\uline{\textcolor{red}{15}},\uline{\textcolor{red}{19}}\}$ & \multirow{2}{*}{$\{\textcolor{red}{12},166\}$}  \\  
\cline{7-7}
               &   $d=-2,e=-1,f=-2\}$  &  $19$  &     &     &    & $\{116\}$ &   \\    
\hline
\multirow{8}{*}{{\tiny TritypeKO} } &  \multirow{8}{*}{$\{i=2,j=3,k=2\}$}  &  \multirow{8}{*}{$54$}  &  \multirow{8}{*}{$\{\textcolor{red}{54}\}$}  &  $\{\uline{26}\}$   &  $\{\uline{26}\}$  &  $\{26\}$ & \multirow{6}{*}{$\{26,27,32,$} \\
\cline{7-7}
\cline{5-6}
&  &  &  & $\{\uline{48}\}$ $\{30\}$,$\{25\}$ & $\{\uline{29},\uline{32}\}$  & $\{\uline{29},\uline{32}\}$ & \multirow{6}{*}{$33,36,48,$} \\
\cline{7-7}
\cline{5-6}
&  &  &  & \multirow{6}{*}{$\{\textcolor{red}{54}\}$} & $\{\uline{48}\}$,$\{30\}$,$\{25\}$ & $\{\uline{29},\uline{35},\uline{57}\}$,$\{25\}$ & \multirow{6}{*}{$57,68\}$} \\
\cline{7-7}
\cline{6-6}
 &    &    &     &     &  $\{\uline{53},\uline{57}\}$,$\{25\}$,$\{30\}$  &  $\{\uline{32},\uline{44},\uline{57}\}$,$\{33\}$,  &  \\
\cline{6-6}
 &    &    &     &     &   \multirow{4}{*}{$\{\textcolor{red}{54}\}$}  & $\{25\}$,$\{30\}$    &   \\ 
\cline{7-7}
 &    &    &     &     &    &  $\{\uline{48}\}$,$\{30\}$,$\{25\}$  &   \\
\cline{7-7}
 &    &    &     &     &    &  $\{\uline{53},\uline{57}\}$,$\{25\}$,$\{30\}$   &   \\
\cline{7-7}
  &    &    &     &     &    &  $\{\textcolor{red}{54}\}$   &   \\     
\hline
\multirow{9}{*}{{\tiny TritypeKO2} } &  \multirow{9}{*}{$\{i=2,j=2,k=4\}$}  &  \multirow{9}{*}{$53$}  &  \multirow{9}{*}{$\{54\}$}  &  $\{\uline{21}\}$   & $\{\uline{21}\}$   &  $\{\uline{21}\}$  &  \multirow{7}{*}{$\{21,26,27,$} \\ 
\cline{6-7}
\cline{5-5}
&    &    &     & $\{\uline{26}\}$   & $\{\uline{26}\}$  &  $\{\uline{26}\}$  & \multirow{7}{*}{$29,30,32,$}  \\
\cline{6-7}
\cline{5-5}
&    &    &     & $\{\uline{35}\}$,$\{27\}$,$\{25\}$ & $\{\uline{29},\uline{57}\}$,$\{30\}$,$\{27\}$,   &  $\{\uline{29},\uline{57}\}$,$\{30\}$,$\{27\}$,  & \multirow{7}{*}{$33,35,36,$}   \\
\cline{5-5}
&    &    &     &   $\{\uline{\textcolor{red}{53}}\}$,$\{25\}$,$\{27\}$   & $\{25\}$   & $\{25\}$  & \multirow{7}{*}{$\textcolor{red}{53},68\}$} \\
\cline{6-7}
\cline{5-5}
&    &    &     &   \multirow{5}{*}{$\{54\}$}  &  $\{\uline{32},\uline{44}\}$,$\{33\}$,$\{25\}$,     & $\{\uline{32},\uline{44}\}$,$\{33\}$,$\{25\}$,    &    \\
&    &    &     &     & $\{27\}$  &  $\{27\}$  &   \\
\cline{6-7}
&    &    &     &     & $\{\uline{35}\}$,$\{27\}$,$\{25\}$  &  $\{\uline{35}\}$,$\{27\}$,$\{25\}$  &   \\
\cline{6-7}
&    &    &     &     &  $\{\uline{\textcolor{red}{53}}\}$,$\{25\}$,$\{27\}$   &  $\{\uline{\textcolor{red}{53}}\}$,$\{25\}$,$\{27\}$  &   \\
\cline{6-7}
&    &    &     &     &  $\{54\}$   &  $\{54\}$  &   \\    
\hline
\multirow{8}{*}{{\tiny TritypeKO2V2} } &  \multirow{8}{*}{$\{i=1,j=2,k=1\}$}   &  \multirow{8}{*}{$31$}  &  \multirow{8}{*}{$\{50\}$}  &  $\{\uline{21}\}$   &  $\{\uline{21}\}$   &  $\{\uline{21}\}$  & \multirow{5}{*}{$\{21,26,27,$} \\
\cline{6-7}
\cline{5-5}
&    &    &     &   $\{\uline{26}\}$  & $\{\uline{26}\}$  &  $\{\uline{26}\}$ & \multirow{5}{*}{$29,\textcolor{red}{31},33,$} \\
\cline{6-7}
\cline{5-5}
&    &    &     &  $\{\uline{29}\}$   & $\{\uline{29}\}$   &  $\{\uline{29}\}$ & \multirow{5}{*}{$34,36,37,$} \\
\cline{6-7}
\cline{5-5} 
&    &    &     &  $\{\uline{36}\}$,$\{\textcolor{red}{31}\}$,$\{25\}$  & $\{\uline{33},\uline{45}\}$,$\{34\}$,$\{\textcolor{red}{31}\}$, &  $\{\uline{33},\uline{45}\}$,$\{34\}$,$\{\textcolor{red}{31}\}$, & \multirow{5}{*}{$49,68\}$} \\ 
\cline{5-5}
&    &    &     &  $\{\uline{49}\}$,$\{\textcolor{red}{31}\}$,$\{25\}$   & $\{25\}$    & $\{25\}$  &  \\ 
\cline{6-7}
\cline{5-5} 
&    &    &     &  \multirow{3}{*}{$\{50\}$}   & $\{\uline{36}\}$,$\{\textcolor{red}{31}\}$,$\{25\}$  &  $\{36\}$,$\{\textcolor{red}{31}\}$,$\{25\}$  &  \\ 
\cline{6-7}
&    &    &     &     &  $\{\uline{49}\}$,$\{\textcolor{red}{31}\}$,$\{25\}$  &  $\{\uline{49}\}$,$\{\textcolor{red}{31}\}$,$\{25\}$ &  \\ 
\cline{6-7}
&    &    &     &     &  $\{50\}$  &  $\{50\}$  &  \\ 
\hline
\multirow{9}{*}{{\tiny TritypeKO3} } &  \multirow{9}{*}{$\{i=1,j=2,k=1\}$}  &  \multirow{9}{*}{$53$}  &  \multirow{9}{*}{$\{54\}$}  &  $\{\uline{21}\}$  & $\{\uline{21}\}$  &  $\{\uline{21}\}$  & \multirow{6}{*}{$\{21,26,27,$} \\
\cline{6-7}
\cline{5-5}
&    &    &     &  $\{\uline{29}\}$   & $\{\uline{26},\uline{57}\}$,$\{30\}$,$\{25\}$,  &  $\{\uline{26},\uline{57}\}$,$\{30\}$,$\{25\}$,  & \multirow{6}{*}{$29,30,32,$} \\
\cline{5-5}
&    &    &     &  $\{\uline{35}\}$,$\{30\}$,$\{25\}$   & $\{27\}$ &  $\{27\}$ & \multirow{6}{*}{$33,35,36,$} \\
\cline{6-7}
\cline{5-5}
&    &    &     &  $\{\uline{\textcolor{red}{53}}\}$,$\{30\}$,$\{25\}$   &  $\{\uline{29}\}$  & $\{\uline{29}\}$  & \multirow{6}{*}{$48,\textcolor{red}{53},68\}$} \\
\cline{6-7}
\cline{5-5} 
&    &    &     &  \multirow{5}{*}{$\{54\}$}   & $\{\uline{32},\uline{44}\}$,$\{33\}$,$\{30\}$,    & $\{\uline{32},\uline{44}\}$,$\{33\}$,$\{30\}$,   &  \\
&    &    &     &     &  $\{25\}$    &  $\{25\}$  &  \\
\cline{6-7}
&    &    &     &     &  $\{\uline{35}\}$,$\{30\}$,$\{25\}$    &  $\{\uline{35}\}$,$\{30\}$,$\{25\}$  &  \\
\cline{6-7}
&    &    &     &     &  $\{\uline{\textcolor{red}{53}}\}$,$\{30\}$,$\{25\}$    &  $\{\uline{\textcolor{red}{53}}\}$,$\{30\}$,$\{25\}$  &  \\
\cline{6-7}
&    &    &     &     &  $\{54\}$    &  $\{54\}$  &  \\
\hline
\multirow{7}{*}{{\tiny TritypeKO4}}  &  \multirow{7}{*}{$\{i=2,j=3,k=3\}$}  &  \multirow{7}{*}{$45$}  &  \multirow{7}{*}{$\{46\}$}  &   $\{\uline{\textcolor{red}{45}}\}$,$\{33\}$,$\{25\}$  &  $\{\uline{26},\uline{32}\}$   &  $\{\uline{26},\uline{32}\}$  &  \multirow{4}{*}{$\{26,27,29,$}  \\
\cline{5-7}
&    &    &     &   \multirow{6}{*}{$\{46\}$}  & $\{\uline{29},\uline{32}\}$   & $\{\uline{29},\uline{32}\}$  & \multirow{4}{*}{$30,32,33,$} \\
\cline{6-7}
&    &    &     &     & $\{\uline{\textcolor{red}{45}}\}$,$\{33\}$,$\{25\}$  &  $\{\uline{32},\uline{35},\uline{49}\}$,$\{25\}$  & \multirow{4}{*}{$35,\textcolor{red}{45},49,$} \\
\cline{6-7}
&    &    &     &     &  \multirow{4}{*}{$\{46\}$}  &  $\{\uline{32},\uline{35},\uline{53}\}$,$\{25\}$  & \multirow{4}{*}{$68\}$} \\
\cline{7-7}
&    &    &     &     &    & $\{\uline{32},\uline{35},\uline{57}\}$,$\{25\}$ &   \\
\cline{7-7}
&    &    &     &     &    & $\{\uline{\textcolor{red}{45}}\}$,$\{33\}$,$\{25\}$   &   \\
\cline{7-7}
&    &    &     &     &    & $\{46\}$  &   \\
\hline
\multirow{7}{*}{{\tiny TritypeKO5}}  &  \multirow{7}{*}{$\{i=2,j=3,k=3\}$}  &  \multirow{7}{*}{$32$,$45$}  &  \multirow{7}{*}{$\{40\}$}  &  $\{\uline{26}\}$   &   $\{\uline{26}\}$ & $\{\uline{26}\}$  & \multirow{5}{*}{$\{26,27,29,$} \\
\cline{5-7}
&    &    &     &  $\{\uline{29}\}$  &  $\{\uline{29}\}$  &  $\{\uline{29}\}$  &  \multirow{5}{*}{$30,\textcolor{red}{32},33,$} \\
\cline{5-7}
&    &    &     &  \multirow{5}{*}{$\{40\}$}   & $\{\uline{\textcolor{red}{32}},\uline{\textcolor{red}{45}}\}$,$\{33\}$,$\{25\}$   & $\{\uline{\textcolor{red}{32}},\uline{\textcolor{red}{45}}\}$,$\{33\}$,$\{25\}$   &  \multirow{5}{*}{$35,49,68\}$} \\
\cline{6-7}
&    &    &     &     & $\{\uline{35},\uline{49}\}$,$\{25\}$  & $\{\uline{35},\uline{49}\}$,$\{25\}$  &   \\
\cline{6-7}
&    &    &     &     & $\{\uline{35},\uline{53}\}$,$\{25\}$  & $\{\uline{35},\uline{53}\}$,$\{25\}$  &   \\  
\cline{6-7}
&    &    &     &     & $\{\uline{35},\uline{57}\}$,$\{25\}$  & $\{\uline{35},\uline{57}\}$,$\{25\}$  &   \\
\cline{6-7}
&    &    &     &     & $\{40\}$   &  $\{40\}$ &   \\  
\hline  
\multirow{9}{*}{{\tiny TritypeKO6} } &  \multirow{9}{*}{$\{i=2,j=3,k=3\}$}  &  \multirow{9}{*}{$32$,$33$}  &  \multirow{9}{*}{$\{40\}$}   &   $\{\uline{26}\}$  &  $\{\uline{26}\}$  & $\{\uline{26}\}$ &  \multirow{7}{*}{$\{26,27,29,$} \\
\cline{5-7}
&    &    &     &  $\{\uline{29}\}$    &  $\{\uline{29}\}$  & $\{\uline{29}\}$  & \multirow{7}{*}{$30,\textcolor{red}{32},\textcolor{red}{33}$} \\ 
\cline{5-7}
&    &    &     &  \multirow{7}{*}{$\{40\}$} &  $\{\uline{35},\uline{49}\}$,$\{25\}$   &   $\{\uline{\textcolor{red}{32}},\uline{45},\uline{49}\}$,$\{\textcolor{red}{33}\}$,$\{25\}$  &  \multirow{7}{*}{$35,49,68\}$} \\
\cline{6-7}
&    &    &     &    & $\{\uline{35},\uline{53}\}$,$\{25\}$ & $\{\uline{\textcolor{red}{32}},\uline{45},\uline{53}\}$,$\{\textcolor{red}{33}\}$,$\{25\}$  &   \\
\cline{6-7}
&    &    &     &    & $\{\uline{35},\uline{57}\}$,$\{25\}$ &  $\{\uline{\textcolor{red}{32}},\uline{45},\uline{57}\}$,$\{\textcolor{red}{33}\}$,$\{25\}$  &   \\  
\cline{6-7}
&    &    &     &    &   \multirow{4}{*}{$\{40\}$} & $\{\uline{35},\uline{49}\}$,$\{25\}$ &   \\
\cline{7-7}
&    &    &     &    &   & $\{\uline{35},\uline{53}\}$,$\{25\}$  &  \\
\cline{7-7}
&    &    &     &    &   & $\{\uline{35},\uline{57}\}$,$\{25\}$  &  \\
\cline{7-7}
&    &    &     &    &   & $\{40\}$   &  \\
\hline
\multirow{3}{*}{{\tiny TriPerimetreKO} } &  \multirow{3}{*}{$\{i=2,j=1,k=2\}$}  &  \multirow{3}{*}{$58$}  &  \multirow{3}{*}{$\{\textcolor{red}{58}\}$}  &  $\{\uline{31}\}$   &  $\{\uline{31}\}$   &  $\{\uline{31}\}$  &  $\{28,29,31,$ \\
\cline{5-7}
&    &    &     &  $\{\uline{37}\}$,$\{32\}$,$\{27\}$   &  $\{\uline{37}\}$,$\{32\}$,$\{27\}$   &  $\{\uline{37}\}$,$\{32\}$,$\{27\}$ & $32,35,37,$  \\
\cline{5-7}
&    &    &     &  $\{\textcolor{red}{58}\}$   & $\{\textcolor{red}{58}\}$   & $\{\textcolor{red}{58}\}$  & $65,72\}$  \\
\hline
\multirow{5}{*}{{\tiny TriPerimetreKOV2} } &  \multirow{5}{*}{$\{i=2,j=3,k=2\}$}  &  \multirow{5}{*}{$34$}  &  \multirow{5}{*}{$\{60\}$,$\{\textcolor{red}{34}\}$}  &  $\{\uline{32}\}$   &    $\{\uline{32}\}$ &  $\{\uline{32}\}$  & $\{28,32,33,$ \\
\cline{5-7}
&    &    &     &  \multirow{2}{*}{$\{\uline{40}\}$,$\{33\}$,$\{27\}$}  & \multirow{2}{*}{$\{\uline{40}\}$,$\{33\}$,$\{27\}$}  &  \multirow{2}{*}{$\{\uline{40}\}$,$\{33\}$,$\{27\}$} &  $\textcolor{red}{34},36,38,$ \\

&    &    &     &  \multirow{4}{*}{$\{60\}$, $\{\textcolor{red}{34}\}$}  &  \multirow{4}{*}{$\{60\}$, $\{\textcolor{red}{34}\}$}  & \multirow{4}{*}{$\{60\}$, $\{\textcolor{red}{34}\}$}  &  $ 40,41,52,$ \\
\cline{5-7}
&    &    &     &     &    &   &  $ 55,56,60,$ \\
&    &    &     &     &    &   &  $ 64,67,74\}$ \\
\hline
\end{tabular}
\end{tiny}
\end{center}
\caption{MCS identifiés  par LocFaults pour des programmes sans boucles}
\label{MCSeskCF}
\end{table*}

\begin{table*}[!Htf]
\begin{center}
\begin{small}
\begin{tabular}{|c|c|c|c|c|c|c|c|}
\hline
\multirow{3}{*}{Programme} & \multicolumn{5}{|c|}{LocFaults} & \multicolumn{2}{c|}{BugAssist} \\
\cline{2-8} &  \multirow{2}{*}{P}  &  \multicolumn{4}{|c|}{L} & \multirow{2}{*}{P} & \multirow{2}{*}{L} \\
\cline{3-6}  &  & $= 0$ & $\leq 1$ & $\leq 2$ & $\leq 3$ &   &    \\
\hline
AbsMinusKO & $0,487s$  & $0,044s$ & $0,073s$ & $0,074s$ & $0,062s$ & $0,02s$ & $0,03s$ \\
\hline
AbsMinusKO2 & $0,484s$  & $0,085s$ & $0,065s$ & $0,085s$ & $0,078s$ & $0,01s$ & $0,06s$ \\
\hline
AbsMinusKO3 & $0,479s$  & $0,076s$ & $ 0,113s$ & $0,357s$ & $0,336s$ & $0,02s$ & $0,04s$ \\
\hline
MinmaxKO & $0,528s$ &  $ 0,243s$ & $0,318s$ & $0,965s$ & $1,016s$ & $0,01s$ & $0,09s$ \\
\hline
MidKO & $0,524s$ & $0,065s$ & $0,078s$ & $0,052s$ & $0,329s$ & $0,02s$ & $0,08s$ \\
\hline
Maxmin6varKO & $0,528s$ & $0,082s$  & $0,132s$ & $0,16s$ & $0,149s$ & $0,06s$ & $1,07s$ \\ 
\hline
Maxmin6varKO2 & $0,536s$ & $0,064s$  & $0,072s$ & $0,097s$ & $0,126s$ & $0,06s$ & $0,66s$ \\ 
\hline
Maxmin6varKO3 & $0,545s$ & $0,066s$ & $0,061s$ & $0,29s$ & $0,307s$ & $0,04s$  & $1,19s$ \\ 
\hline
Maxmin6varKO4 & $0,538s$ & $0,06s$ & $0,07s$ & $0,075s$ & $0,56s$ & $0,04s$ & $0,78s$ \\ 
\hline
TritypeKO & $0,493s$ & $0,022s$  & $0,097s$ & $0,276s$ & $2,139s$ & $0,03s$ & $0,35s$ \\
\hline
TritypeKO2  & $0,51s$ & $0,023s$  & $0,25s$ & $2,083$ & $3,864s$ & $0,02s$ & $0,69s$ \\
\hline 
TritypeKO2V2 & $0,514s$ & $0,034s$ & $0,28s$ & $1,178s$ & $1,31s$ & $0,02s$ & $0,77s$ \\
\hline
TritypeKO3 & $0,493s$ & $0,022s$ & $0,26s$ & $1,928s$ & $4,535s$ & $0,02$ & $0,48s$ \\
\hline
TritypeKO4 & $0,497s$ & $0,023s$ & $0,095s$ & $0,295$ & $5,127s$ & $0,02s$ & $0,21s$ \\
\hline
TritypeKO5 & $0,492s$ & $0,021s$ & $0,099s$ & $0,787s$ & $0,8s$ & $0,01s$ & $0,25s$ \\
\hline
TritypeKO6 & $0,492s$ & $0,025s$ & $0,078s$ & $0,283s$ & $1,841s$ & $0,03s$ & $0,24s$ \\ 
\hline
TriPerimetreKO & $0,518s$ & $0,047s$  & $0,126s$ & $1,096s$ & $2,389s$ & $0,03s$ & $0,64s$ \\
\hline
TriPerimetreKOV2 & $0,503s$ & $0,043s$ & $0,271s$ & $0,639s$ & $1,958s$ & $0,03s$ & $1,20s$ \\
\hline
\end{tabular}
\end{small}
\end{center}
\caption{Temps de calcul}
\label{time1}
\end{table*}

Pour évaluer la méthode que nous avons proposée, nous avons comparé les performances  de {\tt LocFaults} et  de {\tt BugAssist} \cite{JoM11a,JoM11b} sur un ensemble de programmes.  Comme {\tt LocFaults} est basé sur {\tt CPBPV}\cite{CRH10} qui travaille sur des programmes Java et que {\tt BugAssist} travaille sur des programmes C, nous avons construit pour chacun des programmes:
\begin {itemize}
\item une version en Java annotée par une spécification JML;
\item une version en  ANSI-C annotée par la même spécification mais en ACSL.
\end{itemize}
 Les deux versions ont les mêmes numéros de ligne et  les mêmes instructions. La précondition est un contre-exemple du programme, et la postcondition correspond au résultat de la version correcte du programme pour les données du contre-exemple. Nous avons considéré qu'au plus trois conditions pouvaient être fausses sur un chemin. Par ailleurs, nous n'avons pas cherché de MCS de cardinalité supérieure à 3.
Les expérimentations ont été effectuées  avec un processeur Intel Core i7-3720QM 2.60 GHz avec 8 GO de RAM.

Nous avons d'abord utilisé un ensemble de programmes académiques de petite taille (entre 15 et 100 lignes). A savoir :
\begin{itemize}
\item {\textbf{AbsMinus}}. Ce programme  prend en entrée deux entiers $i$ et $j$ et renvoie la valeur absolue de $i-j$. 
\item {\textbf{Minmax}}. Ce programme  prend en entrée trois entiers: $in1$, $in2$ et $in3$, et permet d'affecter la plus petite valeur à la variable $least$ et la plus grande valeur à la variable $most$. 
\item {\textbf{Tritype}}. Ce programme  est un programme classique qui a été utilisé très souvent en test et vérification de programmes. Il prend en entrée trois entiers (les côtés d'un triangle) et retourne $3$ si les entrées correspondent à un triangle équilatéral, $2$ si elles correspondent à un triangle isocèle, $1$ si elles correspondent à un autre type de triangle, $4$ si elles ne correspondent pas à un triangle valide. 
\item {\textbf{TriPerimetre}}.  Ce programme  a exactement la même structure de contrôle que tritype. La différence est que TriPerimetre renvoie la somme des côtés du triangle si les entrées correspondent à un triangle valide, et -1 dans le cas inverse. 
\end{itemize} 
Pour chacun de ces programmes, nous avons considéré différentes versions érronées.

Nous avons aussi évalué notre approche sur les programmes TCAS (Traffic Collision Avoidance System) de la suite de test Siemens\cite{RoW97}. Il s'agit là aussi d'un benchmark bien connu qui correspond à un système d'alerte de trafic et d'évitement de collisions aériennes. Il y a 41 versions erronées et 1608 cas de tests. Nous  avons utilisé toutes les  versions erronées  sauf celles dont l'indice $AltLayerValue$ déborde du tableau $PositiveRAAltThresh$ car les débordements de tableau ne sont pas traités dans {\tt CPBPV}. A savoir, les versions \textbf{TcasKO...TcasKO41}. 
Les erreurs dans ces programmes sont provoquées dans des endroits différents. 1608 cas de tests sont proposés, chacun correspondant à contre-exemple. Pour chacun de ces cas de test $T_{j}$, on construit un programme $TcasV_{i}T_{j}$ qui prend comme entrée le contre-exemple, et dont postcondition correspond à la sortie correcte attendue.

Le code source de l'ensemble des programmes est disponible à l'adresse \url{http://www.i3s.unice.fr/~bekkouch/Bench_Mohammed.html}.

La table ~\ref{MCSeskCF} contient les résultats pour le premier ensemble de programmes:
\begin {itemize}
\item Pour {\tt LocFaults} nous affichons la liste des MCS. La première ligne correspond aux MCS  identifiés sur le chemin initial. Les lignes suivantes aux MCS identifiés sur les chemins pour lesquels la postcondition est satisfaite lorsqu'une condition est déviée. Le numéro de la ligne correspondant à la condition est souligné.\
\item Pour {\tt BugAssist} les résultats correspondent à la fusion de  l'ensemble des compléments des MSS  calculés, fusion qui est opérée par {\tt BugAssist} avant l'affichage des résultats.
\end {itemize}
Sur ces benchmarks les résultats de {\tt LocFaults} sont plus concis et plus précis que ceux de {\tt BugAssist}.

La table \ref{time1} fournit les temps de calcul : dans les deux cas, $P$ correspond au temps de prétraitement  et $L$ au temps de calcul des MCS. Pour {\tt LocFaults}, le temps de pré-traitement inclut la traduction du programme Java en un arbre de syntaxe abstraite avec l'outil JDT (Eclipse Java development tools), ainsi que la construction du CFG dont les noeuds sont des ensembles de contraintes. C'est la traduction Java qui est la plus longue. Pour {\tt BugAssist}, le temps de prétraitement est celui la construction de la formule SAT.
Globalement, les performances de {\tt LocFaults} et {\tt BugAssist} sont similaires bien que le processus d'évaluation de nos systèmes de contraintes soit loin d'être  optimisé. 

La table \ref{TCAS} donne les résultats pour les programmes de la suite {\tt TCAS}. La colonne $Nb\_E$ indique pour chaque programme le nombre d'erreurs qui ont été introduites dans le programme alors que la colonne $Nb\_CE$ donne le nombre de contre-exemples.  Les colonnes $LF$ et $BA$ indiquent respectivement le nombre de contre-exemples pour lesquels  {\tt LocFaults} et {\tt BugAssist} ont identifié l'instruction erronée. On remarquera que {\tt LocFaults} se compare favorablement à  {\tt BugAssist} sur ce benchmark qui ne contient quasiment aucune instruction arithmétique; comme précédemment le nombre d'instructions suspectes identifiées par {\tt LocFaults} est dans l'ensemble nettement inférieur à celui de {\tt BugAssist}.

Les temps de calcul de {\tt BugAssist} et {\tt LocFaults} sont très similaires et inférieurs à une seconde pour chacun des benchmarks de la suite {\tt TCAS}.

\begin{table}[!Htf]
\begin{center}
\begin{scriptsize}
\begin{tabular}{|c|c|c||c||c||}
\hline
Programme & Nb\_E & Nb\_CE & LF & BA \\
\hline
TcasKO & 1 &  131 & 131 & 131 \\
\hline
TcasKO2 & 2 & 67 & 67 &  67 \\
\hline
TcasKO3 & 1 & 23 & 2  &  23 \\
\hline
TcasKO4 & 1 & 20 & 16 & 20 \\
\hline
TcasKO5 & 1  & 10 & 10 & 10 \\
\hline
TcasKO6 & 3  & 12 & 36 & 24 \\ 
\hline
TcasKO7 & 1 & 36 & 23 & 0 \\
\hline
TcasKO8 & 1 & 1  & 1  & 0 \\
\hline
TcasKO9 & 1 & 7 & 7 & 7 \\
\hline
TcasKO10 & 6 & 14 & 16 & 84 \\
\hline
TcasKO11 & 6 & 14 & 16 & 46 \\
\hline
TcasKO12 & 1 & 70 & 52 & 70 \\
\hline
TcasKO13 & 1 & 4 & 3 & 4 \\
\hline
TcasKO14 & 1 & 50 & 6 & 50 \\
\hline
TcasKO16 & 1 & 70 & 22 & 0 \\
\hline
TcasKO17 & 1 & 35 & 22 & 0 \\
\hline
TcasKO18 & 1 & 29 & 21 & 0 \\ 
\hline
TcasKO19 & 1 & 19 & 13 & 0 \\
\hline
TcasKO20 & 1 & 18 & 18 & 18 \\
\hline
TcasKO21 & 1 & 16 & 16 & 16 \\
\hline
TcasKO22 & 1 & 11 & 11 & 11 \\
\hline
TcasKO23 & 1 & 41 & 41 & 41 \\
\hline
TcasKO24 & 1 & 7 & 7 & 7 \\
\hline
TcasKO25 & 1 & 3 & 0 & 3 \\
\hline
TcasKO26 & 1 & 11 & 11 & 11 \\
\hline
TcasKO27 & 1 & 10 & 10 & 10 \\ 
\hline
TcasKO28 & 2 & 75 & 74 & 121 \\ 
\hline
TcasKO29 & 2 & 18 & 17 & 0 \\
\hline
TcasKO30 & 2 & 57 & 57 & 0 \\ 
\hline
TcasKO34 & 1 & 77 & 77 & 77 \\
\hline
TcasKO35 & 4 & 75 & 74 & 115 \\  
\hline
TcasKO36 & 1 & 122 & 120 & 0 \\
\hline
TcasKO37 & 4 & 94 & 110 & 236 \\
\hline
TcasKO39 & 1 & 3 & 0 & 3 \\
\hline
TcasKO40 & 2 & 122 & 0 & 120 \\
\hline
TcasKO41 & 1 & 20 & 17 & 20 \\
\hline
\end{tabular}
\end{scriptsize}
\end{center}
\caption{Nombre  d'erreurs localisés pour TCAS}
\label{TCAS}
\end{table}

\section{Discussion}
Nous avons présenté dans cet article une nouvelle approche pour l'aide à la localisation d'erreurs qui utilise quelques  spécificités de  la programmation par contraintes. Les premiers résultats sont encourageants mais doivent encore être confirmés sur des programmes plus importants et contenant plus d'opérations arithmétiques.

Au niveau des résultats obtenus {\tt LocFaults} est plus précis que {\tt BugAssist} lorsque les erreurs sont sur le chemin du contre exemple ou dans une des conditions du chemin du contre-exemple.  Ceci provient du fait que {\tt BugAssist} et {\tt LocFaults} ne calculent pas exactement la même chose:

{\tt BugAssist} calcule les compléments des différents sous-ensembles obtenus par {\tt MaxSat,} c'est à dire des sous-ensembles de clauses satisfiables de cardinalité maximale. Certaines "erreurs " du programme ne vont pas être identifiées par {\tt BugAssist} car les contraintes correspondantes ne figurent pas dans le complément d'un sous ensemble de clauses satisfiables de cardinalité maximale.

{\tt LocFaults} calcule des MCS, c'est à dire le complément d'un sous-ensemble de clauses maximal, c'est à dire auquel on ne peut pas  ajouter d'autre clause sans le rendre inconsistant, mais qui n'est pas nécessairement de cardinalité maximale. 

{\tt BugAssist}  identifie des instructions suspectes dans l'ensemble du programme alors que {\tt LocFaults} recherche les instructions suspectes sur un seul chemin.

Les travaux futurs concernent à la fois une réflexion sur le traitement des boucles (dans un cadre de bounded-model checking) et l'optimisation de la résolution pour les contraintes numériques. On utilisera aussi  les MCS pour calculer d'autres informations, comme le MUS qui apportent une information complémentaire à l'utilisateur.

\section*{Remerciements:}

Nous tenons à remercier Hiroshi Hosobe, Yahia Lebbah, Si-Mohamed Lamraoui et Shin Nakajima pour les échanges fructueux que nous avons eus.  Nous tenons aussi à remercier  Olivier Ponsini pour son aide et ses précieux conseils lors de la réalisation du prototype de notre système.

\bibliography{mcs_jfpc14}

\end{document}